\documentclass{article}

\usepackage{PRIMEarxiv}

\usepackage[utf8]{inputenc} 
\usepackage[T1]{fontenc}    
\usepackage{hyperref}       
\usepackage{url}            
\usepackage{booktabs}       
\usepackage{nicefrac}       
\usepackage{microtype}      
\usepackage{fancyhdr}       
\usepackage{graphicx}       

\usepackage{subcaption}
\usepackage[nolist]{acronym}
\usepackage{tabularx}
\usepackage{longtable}
\usepackage{listings}
\usepackage{multirow}

\usepackage{amsmath,amssymb,amsfonts}%
\usepackage{amsthm}%
\usepackage{mathrsfs}%
\usepackage{xcolor}%
\usepackage{textcomp}%
\usepackage{manyfoot}%

\usepackage[normalem]{ulem}
\usepackage{authblk}
\useunder{\uline}{\ul}{}

\title{Augmenting representations with scientific papers}

\author[1, 2, 3, *]{Nicolò Oreste Pinciroli Vago}
\author[3]{Rocco Di Tella}
\author[3, 4]{Carolina Cuesta-Lázaro}
\author[3]{Michael J. Smith}
\author[3]{Cecilia Garraffo}
\author[3]{Rafael Mart\'{\i}nez-Galarza}

\affil[1]{Department of Electronics, Information and Bioengineering, Politecnico di Milano \\ 
Via Giuseppe Ponzio, 34, Milan, 20133, MI, Italy}

\affil[2]{Osservatorio Astronomico di Roma, INAF \\ 
Via Frascati 33, Monte Porzio Catone, 00078, RM, Italy}

\affil[3]{AstroAI, Center for Astrophysics $|$ Harvard \& Smithsonian \\ 
60 Garden Street, Cambridge, 02138, MA, United States of America}

\affil[4]{Center for Computational Astrophysics, Institute for Advanced Study/The Flatiron Institute \\ 
162 5th Ave 9th floor, New York, 10010, NY, United States of America}

\affil[*]{\normalfont \texttt{nicolooreste.pinciroli@polimi.it}}

\begin{document}
\maketitle

\begin{abstract}
Astronomers have acquired vast repositories of multimodal data, including images, spectra, and time series, complemented by decades of literature that analyzes astrophysical sources. Still, these data sources are rarely systematically integrated. This work introduces a contrastive learning framework designed to align X-ray spectra with domain knowledge extracted from scientific literature, facilitating the development of shared multimodal representations. Establishing this connection is inherently complex, as scientific texts encompass a broader and more diverse physical context than spectra. We propose a contrastive pipeline that achieves a 20\% Recall@1\% when retrieving texts from spectra, proving that a meaningful alignment between these  modalities is not only possible but capable of accelerating the interpretation of rare or poorly understood sources. Furthermore, the resulting shared latent space effectively encodes physically significant information. By fusing spectral and textual data, we improve the estimation of 20 physical variables by $16-18\%$ over unimodal spectral baselines. Our results indicate that a Mixture of Experts (MoE) strategy, which leverages both unimodal and shared representations, yields superior performance. Finally, outlier analysis within the multimodal latent space identifies high-priority targets for follow-up investigation, including a candidate pulsating ULX (PULX) and a gravitational lens system. Importantly, this framework can be extended to other scientific domains where aligning observational data with existing literature is possible.
\end{abstract}

\section{Introduction}

Foundation models are large-scale neural networks pre-trained on diverse data and adaptable to multiple downstream tasks, and have recently been applied to astronomy \cite{Parker24, leung_astronomical_2024}. In astronomy, upcoming surveys (Vera Rubin Observatory, Roman Space Telescope) will generate petabyte-scale multimodal datasets \cite{greenstreet_impact_2024, Hernandez24, Gezari22}, which need scalable approaches to extract scientific insights. Unlike single-modality foundation models, astronomical observations are inherently multimodal: a single source may have images, spectra, light curves, and decades of textual descriptions in scientific literature, each capturing complementary physical information.

There exist both unimodal and multimodal astronomical foundation models \cite{Parker24, Mishra-Sharma24, rizhko_astrom3_2025}. However, the systematic integration of observational data with textual scientific knowledge remains unexplored. This gap is significant, as scientific literature contains high-quality peer-reviewed expert interpretations, physical models, and contextual information unavailable in raw observations alone.

We present the first contrastive learning framework aligning X-ray spectra with scientific papers' summaries, creating a shared latent space that enhances spectral data and encodes physical properties. This approach addresses key challenges for astronomical foundation models by fusing heterogeneous data modalities such as spectra and text, preserving physically meaningful structures in learned representations, and enabling knowledge transfer across observational datasets and scientific literature.

The contributions of this work can be summarized as follows: (1) alignment of X-ray spectra with textual summaries using contrastive learning; (2) demonstration that multimodal representations outperform unimodal ones for physical parameter estimation (3) 97\% data compression (4,672 to 128 dimensions) while preserving relevant physical information; and (4) ability to use the enriched latent space to flag outliers.

\section{Methods}

\subsection{Dataset}

An X-ray spectrum is a measurement of the distribution of detected photons across a range of energy levels. For Chandra observations \cite{Evans24}, a spectrum is derived from individual photon events between 0.5 and 8 keV. In this work, for each source (extracted from the Chandra Source Catalog), the energy range is discretized into 400 bins where the intensity is measured as a photon count rate (photons per unit of time and energy). These data are min-max normalized to allow the model to learn from the relative distribution of energy (the spectral shape), which serves as a physical signature of the source's underlying processes (see also \cite{vago_extracting_2025}).

To incorporate expert knowledge, we cross-reference the spectral data with the NASA Astrophysics Data System (ADS), which contains a collection of scientific papers in astronomy and astrophysics. Cross-referencing is allowed by the use of sky coordinates and source identifiers from SIMBAD. 

The final dataset consists of 11,447 spectrum-text pairs, split into training (69\%), calibration (1\%), validation (15\%), and test (15\%) sets. Each sample is associated with up to 20 ground-truth physical variables (see Table~\ref{tab:variables}) used to evaluate the physical consistency of the learned representations.

\begin{table}
\centering
\caption{Physical variables and their descriptions from the Chandra Source Catalog. These 20 variables serve as the ground truth for evaluating the physical interpretability of the learned latent representations and are used in the multimodal regression tasks.}
\label{tab:variables}
\begin{tabularx}{\textwidth}{@{}lX@{}}
\toprule
\textbf{Variable} & \textbf{Description} \\ \midrule
\texttt{hard\_hs} & Hardness ratio between hard (2.0--7.0 keV) and soft (0.5--1.2 keV) bands \\
\texttt{hard\_ms} & Hardness ratio between medium (1.2--2.0 keV) and soft (0.5--1.2 keV) bands \\
\texttt{hard\_hm} & Hardness ratio between hard (2.0--7.0 keV) and medium (1.2--2.0 keV) bands \\
\texttt{var\_prob\_b} & Gregory–Loredo variability probability \\
\texttt{var\_index\_b} & Intra-observation Gregory-Loredo variability index \\
\texttt{powlaw\_gamma} & Photon index from power-law spectral fit \\
\texttt{powlaw\_nh} & Hydrogen column density from power-law fit \\
\texttt{powlaw\_stat} & Fit statistic for the power-law spectral model \\
\texttt{bb\_kt} & Temperature (keV) from blackbody spectral fit \\
\texttt{bb\_nh} & Hydrogen column density from blackbody fit \\
\texttt{bb\_stat} & Fit statistic for the blackbody model \\
\texttt{brems\_kt} & Temperature (keV) from Bremsstrahlung fit \\
\texttt{brems\_nh} & Hydrogen column density from bremsstrahlung fit \\
\texttt{brems\_stat} & Fit statistic for the bremsstrahlung model \\
\texttt{apec\_kt} & Temperature (keV) in \ac{APEC} thermal model fit \\
\texttt{apec\_abund} & Abundance of the best fitting absorbed \ac{APEC} model spectrum to the source region aperture Pulse Invariant spectrum \\
\texttt{apec\_z} & Redshift in \ac{APEC} fit \\
\texttt{apec\_nh} & Hydrogen column density from \ac{APEC} fit \\
\texttt{apec\_stat} & Fit statistic for the \ac{APEC} model \\
\texttt{flux\_significance\_b} & Flux significance \\ 
\bottomrule
\end{tabularx}
\end{table}
\subsection{Architecture}

\begin{figure}[h]
\centering
\includegraphics[width=\textwidth]{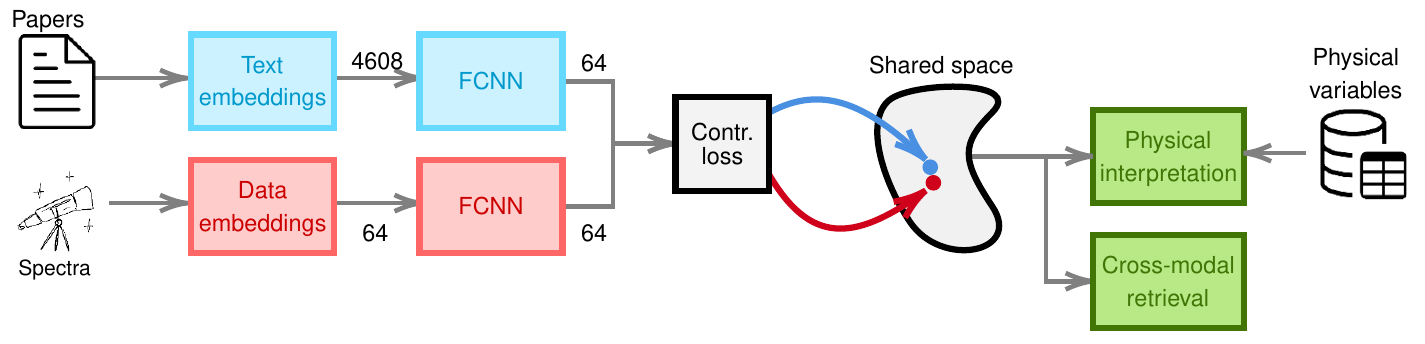}
\caption{Pipeline overview. Spectra are encoded via a transformer-based autoencoder. 
Scientific papers are summarized using GPT-4o-mini, and the summaries are embedded from OpenAI's Ada-002. Contrastive learning aligns modalities into a shared latent space for downstream tasks.}
\label{fig:pipeline}
\end{figure}

Our pipeline, presented in Figure~\ref{fig:pipeline}, follows foundation model principles: we use two pre-trained unimodal foundation models followed by contrastive alignment, focusing the evaluation on downstream tasks.

The spectra are processed by the transformer-based autoencoder introduced in \cite{vago_extracting_2025}. In this work, we compress spectra to 64-dimensional latent vectors, minimizing the reconstruction loss (MAE). The textual summaries are generated from scientific papers using GPT-4o-mini and later embedded using OpenAI's Ada-002 model.

Two fully-connected networks map spectral (64-dimensional) and text (4,608-dimensional) embeddings to a shared 64-dimensional space. We optimize the InfoNCE loss \cite{oord_representation_2019}:
\begin{equation}
\mathcal L_{\text{InfoNCE}} = -\frac{1}{N}\sum_{i=1}^{N} \log \frac{\exp(\text{sim}(t_i, d_i)/\tau)}{\sum_{j=1}^{N} \exp(\text{sim}(t_i, d_j)/\tau)}
\end{equation}
where $\text{sim}(x,y)$ is cosine similarity, $\tau$ is temperature (tuned on calibration set), and $(t_i, d_i)$ are matched text-data pairs.

We also evaluate the results on three downstream tasks:
\begin{itemize}
    \item Cross-modal retrieval: we retrieve text descriptions from spectra using similarity search.
    \item Physical parameter regression: we use a $k$-NN regressor (with $k=3$) to predict 20 physical variables from latent representations. We employ a Mixture of Experts (MoE) strategy: for each variable, we select the best representation (pre- or post-alignment, using texts, spectra, or both) based on validation set Pearson correlation.
    \item Outlier detection: we use Isolation Forest to identify rare sources in the aligned latent space.
\end{itemize} 

\subsection{Training and evaluation}

Models are trained with Adam optimizer, performing a grid search in the hyperparameter space over learning rate ($10^{-4}$ to $10^{-3}$), shared space dimension (16 to 128), dropout (0.1 to 0.5), and hidden dimensions (16 to 1024). The performance metrics are Recall@k\% (i.e., the proportion of queries with correct match in top $k$\%), Median Rank, MAE (for regression), and Pearson correlation (for latent space-physical variables relationships).

\section{Results}

\subsection{Cross-modal retrieval}

\begin{figure}
    \centering
    \includegraphics[width=\linewidth]{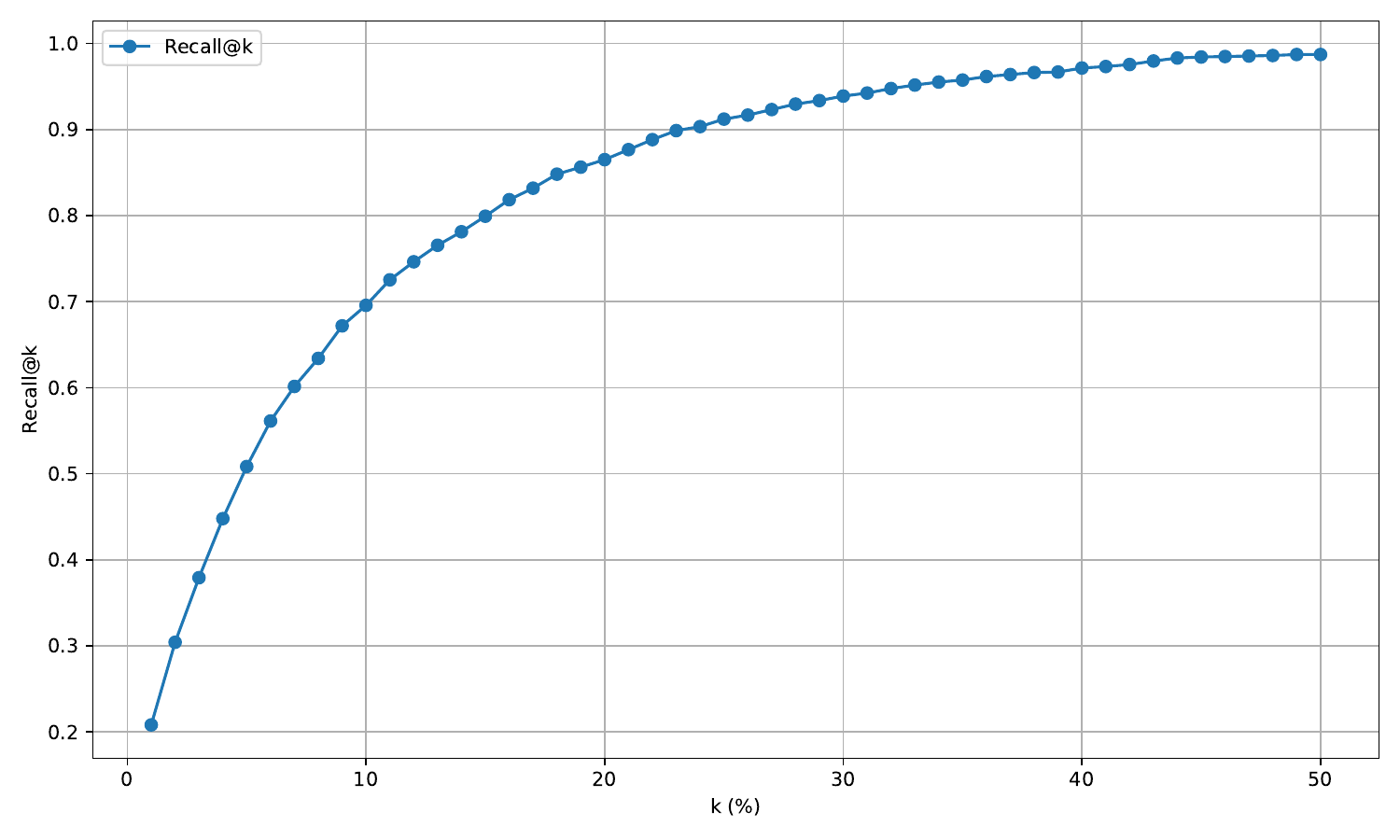}
    \caption{Recall@k\% as a function of $k$, expressed as a percentage of the test set, for the ensemble model.}
    \label{fig:evolution_recall}
\end{figure}

Our pipeline achieves $\approx 20\%$ Recall@1\%  and $\approx$ 50\% Recall@5\% (corresponding to a Median Rank of 84). This indicates the median distance of each spectrum to the matching scientific summary is 84 among 1,719 candidates, exploring $\approx 5\%$ of the search space. Figure \ref{fig:evolution_recall} presents the Recall@$k$\% for varying values of $k$.

\subsection{Physical interpretation}

Unsurprisingly, autoencoders on spectra are better correlated with the selected physical variables (on average, $|\rho| = 0.43$) than the autoencoders on texts (on average, $|\rho| = 0.30$). Aligning and combining both modalities leads to an even stronger alignment (average $|\rho| = 0.55$). Table~\ref{tab:correlations} shows top correlations: for instance, dimensions $L_{12}$ and $L_1$  encode hardness ratio (hard\_hs, $\rho = 0.82$), while $L_{48}$ a captures thermal properties (apec\_kt, $\rho \approx 0.74$).

\begin{table}[h]
\centering
\caption{Top 10 latent-variable correlations (post-alignment spectra embeddings).}
\label{tab:correlations}
\begin{tabular}{lll}
\toprule
Latent Dim. & Variable & Correlation \\
\midrule
$L_{12}$ & hard\_hs & 0.82 \\
$L_1$ & hard\_hs & 0.82 \\
$L_{51}$ & hard\_hs & 0.77 \\
$L_{26}$ & hard\_hs & 0.76 \\
$L_{48}$ & apec\_kt & 0.74 \\
$L_{29}$ & hard\_hs & 0.74 \\
$L_{44}$ & hard\_hs & 0.71 \\
$L_8$ & powlaw\_gamma & 0.68 \\
$L_{30}$ & brems\_nh & 0.68 \\
$L_{62}$ & bb\_kt & 0.68 \\
\bottomrule
\end{tabular}
\end{table}

The results also indicate that alignment with scientific papers enhances physical interpretability: the shared latent space better reflects physical variables than either modality alone.

\begin{figure}[htbp]
    \centering

    \begin{subfigure}[b]{0.3\textwidth}
        \includegraphics[width=\linewidth]{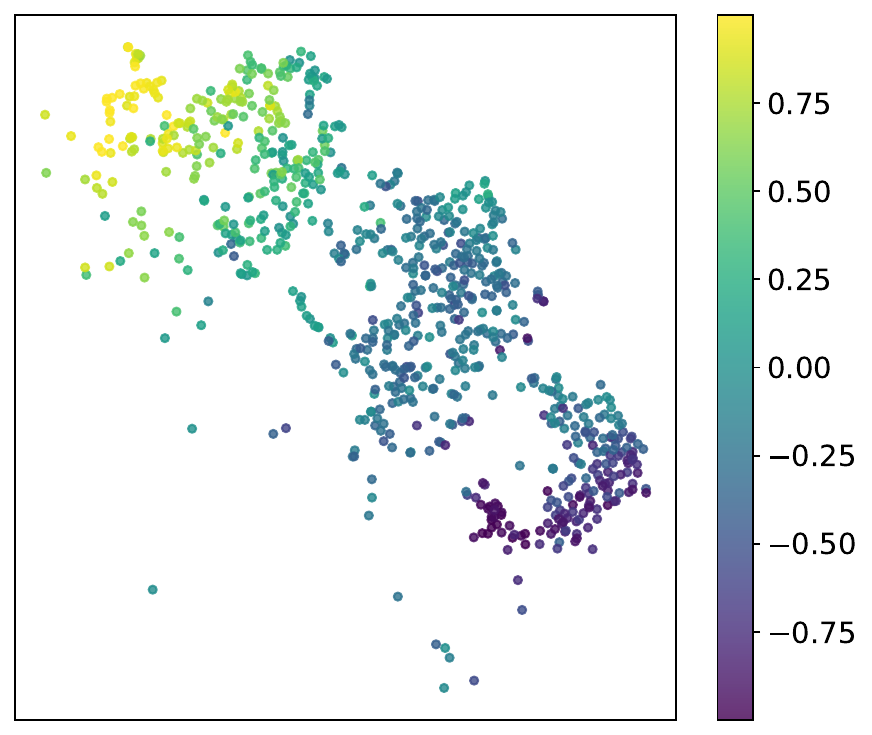}
        \caption{hard\_hs}
    \end{subfigure}
    \begin{subfigure}[b]{0.3\textwidth}
        \includegraphics[width=\linewidth]{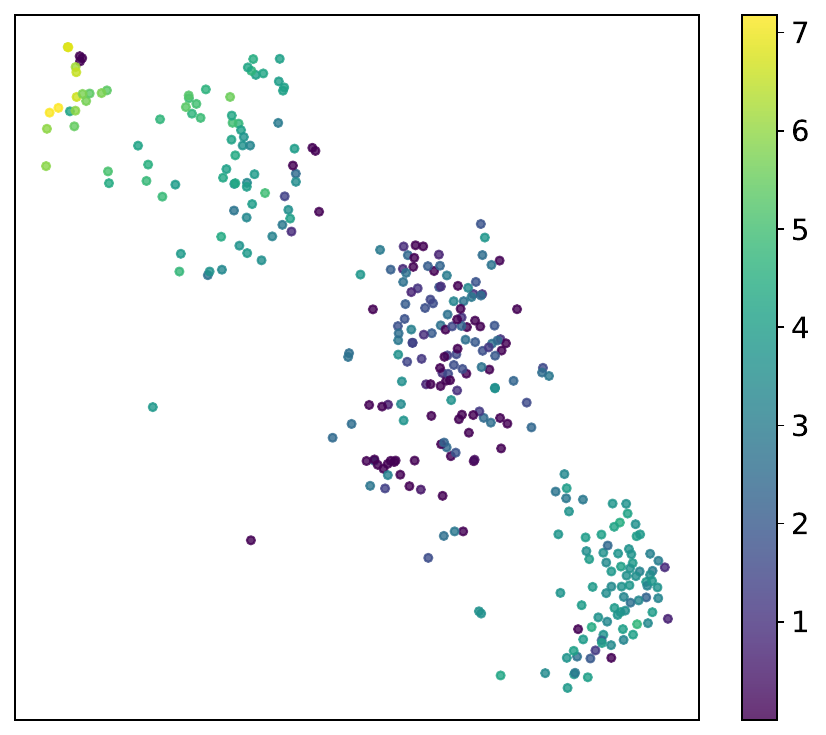}
        \caption{powlaw\_nh (log scale)}
    \end{subfigure}
    \begin{subfigure}[b]{0.3\textwidth}
        \includegraphics[width=\linewidth]{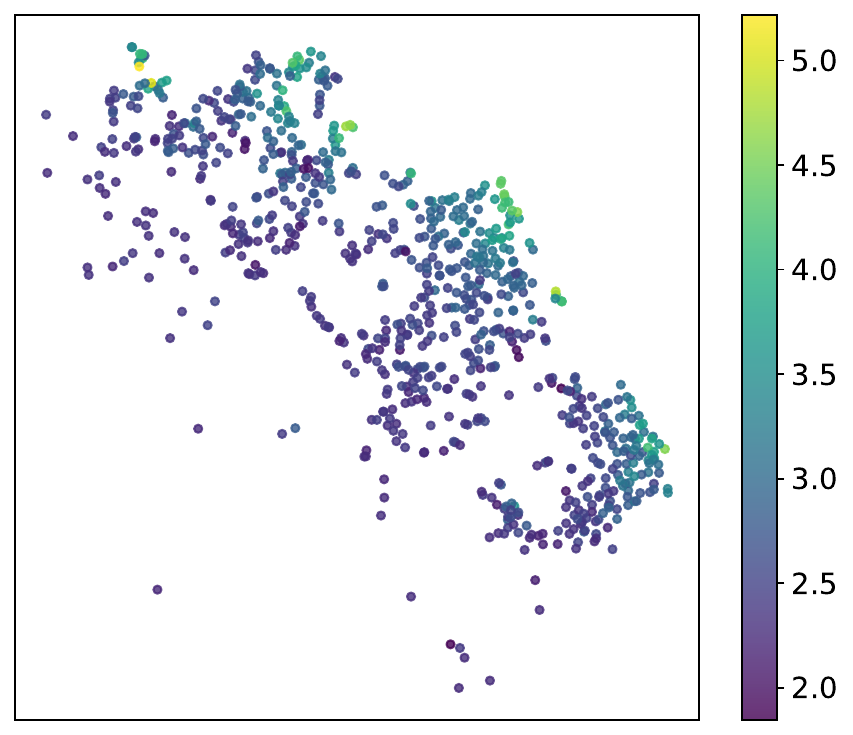}
        \caption{flux\_significance\_b (log scale)}
    \end{subfigure}

    \caption{tSNE plots for a subset of the physical variables}
    \label{fig:tsne}
\end{figure}

Beyond individual linear correlations, we analyze whether the latent space is well-structured with respect to the 20 physical variables. To this end, we use $k$-NN to estimate them. Figure \ref{fig:tsne} presents three tSNE plots that show that multimodal representations generate a well-structured space with respect to a subset of the physical variables. Moreover, multimodal representations substantially improve parameter estimation (Table~\ref{tab:regression}). Using both aligned modalities reduces MAE by $\approx 16\%$ with respect to the best pre-alignment modality, and the MoE strategy further improves to $\approx 18\%$. For hardness ratios (hard\_hs, hard\_ms, hard\_hm), average improvement is 34\%. Hydrogen column density ($N_H$) estimates improve by 34\% across spectral models. For variability metrics (var\_prob\_b, var\_index\_b), text alone performs better because spectral data lacks temporal information, which is lost during alignment. 

\begin{table}[]
\centering
\caption{Comparison of \ac{MAE} before and after alignment, obtained using a $k$-NN ($k=3$) regressor. The ``Mean baseline'' column reports the \ac{MAE} obtained when predicting the validation set mean for all samples. The ``Improvement'' column indicates the relative improvement of the \ac{MoE} strategy compared to the best-performing pre-alignment modality. Underlined values represent the lowest \ac{MAE} (best performance), while bold values indicate results within the confidence interval of the best value.}\label{tab:regression}
\resizebox{\columnwidth}{!}{%
\begin{tabular}{@{}lc @{\hspace{1em}} cc @{\hspace{1em}} ccc @{\hspace{1em}} cc @{\hspace{1em}} c@{}}
\toprule
\multirow{2}{*}{\textbf{Variable}} & \multirow{2}{*}{\textbf{Mean baseline}} & \multicolumn{2}{c}{\textbf{Pre-alignment}} & \multicolumn{3}{c}{\textbf{Post-alignment}} & \multirow{2}{*}{\textbf{\ac{MoE}}} & \multirow{2}{*}{\textbf{Uncertainty}} & \multirow{2}{*}{\textbf{Improvement}} \\
\cmidrule(lr){3-4} \cmidrule(lr){5-7}
 &  & \textbf{Spectra} & \textbf{Text} & \textbf{Spectra} & \textbf{Text} & \textbf{Both} &  &  &  \\ 
\midrule
\texttt{hard\_hs} & 0.40 & 0.20 & 0.27 & 0.16 & 0.17 & {\ul \textbf{0.12}} & {\ul \textbf{0.12}} & 0.01 & 40\% \\
\texttt{hard\_ms} & 0.28 & 0.17 & 0.22 & 0.14 & 0.17 & {\ul \textbf{0.12}} & {\ul \textbf{0.12}} & 0.01 & 28\% \\
\texttt{hard\_hm} & 0.23 & 0.15 & 0.18 & 0.11 & 0.15 & {\ul \textbf{0.10}} & {\ul \textbf{0.10}} & 0.01 & 33\% \\
\texttt{var\_prob\_b} & 0.26 & 0.27 & {\ul \textbf{0.18}} & 0.26 & {\ul 0.20} & 0.23 & 0.23 & 0.02 & $-25$\% \\
\texttt{var\_index\_b} & 1.77 & 1.79 & {\ul \textbf{1.00}} & 1.70 & 1.19 & 1.45 & {\ul \textbf{1.00}} & 0.10 & 0\% \\
\texttt{powlaw\_gamma} & 0.89 & 0.67 & 0.65 & {\ul 0.44} & 0.67 & {\ul \textbf{0.41}} & {\ul \textbf{0.41}} & 0.05 & 36\% \\
\texttt{powlaw\_nh} & 65.32 & 33.08 & 48.88 & 26.39 & 40.82 & {\ul \textbf{21.63}} & {\ul \textbf{21.63}} & 2.71 & 35\% \\
\texttt{powlaw\_stat} & 0.47 & 0.46 & 0.43 & {\ul \textbf{0.36}} & 0.52 & {\ul 0.37} & {\ul 0.37} & 0.04 & 14\% \\
\texttt{bb\_kt} & 3.23 & 0.32 & 0.49 & {\ul \textbf{0.27}} & 0.40 & {\ul 0.28} & {\ul 0.28} & 0.03 & 13\% \\
\texttt{bb\_nh} & 58.95 & 74.31 & 111.48 & {\ul 53.78} & 101.59 & {\ul \textbf{50.29}} & {\ul \textbf{50.29}} & 6.30 & 32\% \\
\texttt{bb\_stat} & 0.91 & 0.75 & 0.78 & {\ul \textbf{0.65}} & 0.79 & {\ul 0.68} & {\ul 0.68} & 0.08 & 10\% \\
\texttt{brems\_kt} & 11.76 & 6.26 & 6.27 & {\ul 5.37} & {\ul \textbf{5.35}} & {\ul 5.64} & {\ul 5.64} & 0.67 & 10\% \\
\texttt{brems\_nh} & 125.98 & 23.31 & 26.98 & {\ul 15.57} & 19.54 & {\ul \textbf{14.12}} & {\ul \textbf{14.12}} & 1.77 & 39\% \\
\texttt{brems\_stat} & 0.57 & 0.53 & 0.50 & {\ul \textbf{0.42}} & 0.62 & {\ul 0.46} & {\ul 0.46} & 0.05 & 9\% \\
\texttt{apec\_kt} & 3.51 & 1.60 & 1.74 & {\ul 1.23} & 1.60 & {\ul \textbf{1.22}} & {\ul \textbf{1.22}} & 0.23 & 23\% \\
\texttt{apec\_abund} & 0.72 & {\ul 0.76} & 0.89 & {\ul \textbf{0.70}} & {\ul 0.78} & {\ul 0.81} & {\ul 0.81} & 0.13 & $-6$\% \\
\texttt{apec\_z} & 0.22 & {\ul 0.21} & {\ul 0.21} & {\ul 0.20} & {\ul 0.21} & {\ul 0.20} & {\ul \textbf{0.21}} & 0.04 & $-3$\% \\
\texttt{apec\_nh} & 65.84 & 35.95 & 35.67 & {\ul 27.77} & 31.47 & {\ul \textbf{24.92}} & {\ul \textbf{24.92}} & 4.65 & 30\% \\
\texttt{apec\_stat} & 0.73 & {\ul 0.66} & 0.75 & {\ul \textbf{0.61}} & 0.78 & {\ul 0.62} & {\ul 0.62} & 0.11 & 6\% \\
\texttt{flux\_significance\_b} & 8.50 & 7.36 & 7.67 & {\ul \textbf{4.06}} & 6.53 & 4.54 & 4.54 & 0.42 & 38\% \\ 
\bottomrule
\end{tabular}%
}
\end{table}

The 97\% compression (4,672 to 128 dimensions\footnote{64 dimensions in the shared space for each modality}) while retaining predictive power is critical for scaling to billion-object surveys (e.g., LSST), where full-dimensionality similarity searches are intractable.

Overall, the emerging structure, not explicitly enforced during training, indicates that contrastive learning preserves domain-relevant information.

\subsection{Outlier detection}

Beyond regression and retrieval, the shared latent space enables the discovery of rare astronomical phenomena by identifying points that deviate from the learned multimodal manifold. We apply Isolation Forest, an unsupervised algorithm that isolates outliers by randomly partitioning the high-dimensional feature space, to the 1,719 objects in the test set. By operating on the aligned latent space, this approach identifies sources where the combination of spectral and textual features is a statistical outlier, potentially flagging objects that challenge standard physical models. The analysis of the test set reveals class-level patterns and high-confidence outliers.

Class-level patterns indicate that Quasars (QSOs) exhibit higher median anomaly scores than typical AGNs, reflecting their extreme luminosities. Ultra-luminous X-ray sources (ULXs) show high variance, consistent with the presence of subpopulations (pulsating vs. non-pulsating). In addition, the top 1\% anomalies include the gravitational lens system 2CXOJ224030.2+032131 and the ULX 2CXOJ004722.6-252050. The latter has been independently identified as a candidate pulsating ULX candidate in \cite{pinciroli_vago_hunt_2025}, validating the pipeline's ability to discover scientifically interesting objects. It is important to note that the findings in \cite{pinciroli_vago_hunt_2025} were not included in our training dataset, as the publication date succeeded our data collection cut-off. Consequently, the identification of this source as an outlier serves as an independent validation of our model's discovery potential.

\section{Discussion}

Beyond astronomy, this approach applies to any domain with paired observational sequences and textual annotations, such as seismology (which includes waveforms and event reports \cite{si_seisclip_2024}), climate science (which includes timeseries and assessment documents \cite{10.1145/3626772.3657950}), and medicine (which comprises physiological signals and clinical notes \cite{liu_multimodal_2023}). Our work demonstrates that scientific literature, which is widely available, easy to process, and represents a vast repository of expert knowledge, can be systematically integrated with observational data to create enriched foundation models. This knowledge-augmented paradigm leverages decades of domain expertise, accelerating interpretation and discovery.

For astronomy, this enables:
\begin{itemize}
\item Semantic search: given a spectrum, retrieve relevant papers and similar sources, facilitating literature exploration.
\item Cross-facility fusion: combine spectra from different instruments (e.g., Chandra, XMM-Newton) via shared textual metadata.
\item Automated characterization: preliminary source classification and parameter estimation from latent space, prioritizing follow-up observations.
\end{itemize}

Moreover, our findings highlight that:

\begin{itemize}
    \item Contrastive learning not only enables cross-modal retrieval but also produces latent spaces more correlated with physical variables, an emergent property valuable for scientific applications.
    \item 97\% compression facilitates billion-scale similarity searches, essential for next-generation surveys. 
\end{itemize}

Still, this work has some limitations:

\begin{itemize}
    \item The retrieval performance of 20\% Recall@1\% suggests room for improvement. This limitation can be mitigated by improving text summaries and leveraging additional data pairs (possibly adding multimodal observations). Still, the mismatch between X-ray spectra and scientific summaries prevents a perfect alignment.
    \item Current work focuses on retrieval and regression, and has not proven its usefulness on tasks like text generation from spectra
    \item Anomaly detection could enhance outlier detection by incorporating physics-based priors, with the goal to prioritize theoretically interesting outliers over statistical artifacts.
    \item The alignment is limited to astrophysical data, even though the same pipeline can be applied to other fields of science
\end{itemize}

\section{Conclusions}

We present the first multimodal foundation model aligning X-ray spectra with scientific texts, demonstrating that knowledge-augmented representation learning can bridge observational data and domain expertise. Key achievements include: (1) 18\% improvement in physical parameter estimation via multimodal fusion; (2) 97\% data compression preserving correlations with 20 physical variables; (3) discovery of rare phenomena (candidate PULXs, gravitational lenses) through latent space outlier detection.

As petabyte-scale astronomical surveys come online, scalable multimodal approaches are necessary. Our framework, combining self-supervised pre-training, contrastive alignment, and ensemble-based downstream adaptation, offers a blueprint for integrating heterogeneous scientific data. Importantly, this framework can be extended to other scientific domains where aligning observational data with existing literature is possible. By systematically connecting observations with textual knowledge, we move toward foundation models that not only process data efficiently but also encode the semantic richness of scientific understanding.

\section*{Data Availability}
Processed data: \url{https://osf.io/56s4h/overview?view_only=1ead08e953814767a6f74e02b73fc40b}.

\begin{acronym}
\acro{AGN}{Active Galactic Nucleus}
\acro{APEC}{Astrophysical Plasma Emission Code}
\acro{CIAO}{Chandra Interactive Analysis of Observations}
\acro{InfoNCE}{Information Noise-Contrastive Estimation}
\acro{ISOF}{Isolation Forest}
\acro{kNN}{$k$-Nearest Neighbor}
\acro{MAE}{Mean Absolute Error}
\acro{MoE}{Mixture of Experts}
\acro{PULX}{Pulsating Ultra-Luminous X-ray source}
\acro{SIMBAD}{Set of Identifications, Measurements and Bibliography for Astronomical Data}
\acro{ULX}{Ultra-Luminous X-ray source}
\acro{YSO}{Young Stellar Object}
\end{acronym}

\section*{Acknowledgments}
This work was conducted as part of the AstroAI collaboration. The authors gratefully acknowledge Piero Fraternali for his careful proofreading. They also wish to thank Melanie Weber, David Alvarez-Melis, and the AstroAI multimodal group at CfA for their valuable suggestions and insightful discussions. The corpus of literature data was curated and prepared by NASA's Astrophysics Data System (ADS), and data from the Chandra Archive and Chandra Source Catalog were used. This research was supported by computational resources provided by the CINECA Leonardo supercomputer and by INAF through the project INA24\_C6B05. RMG was supported by Astromind, Inc. NOPV and RMG were supported by AstroAI.

\bibliographystyle{unsrt}  
\bibliography{references}

\end{document}